\documentclass[a4paper, 10pt, conference]{ieeeconf}      % Use this line for a4
\usepackage{FG2026}
\usepackage{amsmath, amssymb, mathrsfs}
\usepackage{cleveref}
\usepackage{graphicx}
\usepackage{amsfonts}
\usepackage{multirow}
\usepackage{algorithm}
\usepackage{algorithmic}
\FGfinalcopy % *** Uncomment this line for the final submission
\usepackage[table]{xcolor} % Add this in your preamble
\definecolor{best}{RGB}{60,179,113} % SkyBlue
\definecolor{second}{RGB}{50,205,50}    % LightSkyBlue
\definecolor{third}{RGB}{176,224,230}    % PowderBlue

\IEEEoverridecommandlockouts                              % This command is only
\def\FGPaperID{45} % *** Enter the FG2026 Paper ID here

\title{\LARGE \bf
Mesh-Gait: A Unified Framework for Gait Recognition Through Multi-Modal Representation Learning from 2D Silhouettes
}

\author{
	Zhao-Yang Wang$^{1}$
        \;\; Jieneng Chen$^{1}$
	\;\; Jiang Liu$^{2}$
	\;\; Yuxiang Guo$^{1}$	
	\;\; Rama Chellappa$^{1}$ \\
	$^1$Johns Hopkins University \;\; 
    $^2$Advanced Micro Devices, Inc. \;\; 
 \\ 
}

\begin{document}

\ifFGfinal
\thispagestyle{empty}
\pagestyle{empty}
\else
\author{Anonymous FG2026 submission\\ Paper ID \FGPaperID \\}
\pagestyle{plain}
\fi
\maketitle

%%%%%%%%%%%%%%%%%%%%%%%%%%%%%%%%%%%%%%%%%%%%%%%%%%%%%%%%%%%%%%%%%%%%%%%%%%%%%%%%
\begin{abstract}

Gait recognition, a fundamental biometric technology, leverages unique walking patterns for individual identification, typically using 2D representations such as silhouettes or skeletons. However, these methods often struggle with viewpoint variations, occlusions, and noise. Multi-modal approaches that incorporate 3D body shape information offer improved robustness but are computationally expensive, limiting their feasibility for real-time applications.

To address these challenges, we introduce Mesh-Gait, a novel end-to-end multi-modal gait recognition framework that directly reconstructs 3D representations from 2D silhouettes, effectively combining the strengths of both modalities. Compared to existing methods, directly learning 3D features from 3D joints or meshes is complex and difficult to fuse with silhouette-based gait features. To overcome this, Mesh-Gait reconstructs 3D heatmaps as an intermediate representation, enabling the model to effectively capture 3D geometric information while maintaining simplicity and computational efficiency. During training, the intermediate 3D heatmaps are gradually reconstructed and become increasingly accurate under supervised learning, where the loss is calculated between the reconstructed 3D joints, virtual markers, and 3D meshes and their corresponding ground truth, ensuring precise spatial alignment and consistent 3D structure. Mesh-Gait extracts discriminative features from both 2D silhouettes and reconstructed 3D heatmaps in a computationally efficient manner. This design enables the model to capture spatial and structural gait characteristics while avoiding the heavy overhead of direct 3D reconstruction from RGB videos, allowing the network to focus on motion dynamics rather than irrelevant visual details.

Extensive experiments on multiple benchmark datasets demonstrate that Mesh-Gait not only generates high-quality 3D gait representations but also achieves state-of-the-art recognition accuracy and robustness. These results highlight the potential of Mesh-Gait for real-world gait recognition applications. The code will be released upon acceptance of the paper.

\end{abstract}

\section{Introduction}
\label{sec:intro}

Gait recognition has emerged as a critical biometric modality, offering a non-intrusive and efficient method for identifying individuals based on their unique walking patterns. Unlike other biometric technologies, such as facial recognition or fingerprint scanning, gait recognition allows identification from a long distance without requiring active cooperation from the individual. This distinctive capability makes it particularly suitable for applications in surveillance~\cite{benedek2016lidar}, security authentication~\cite{shen2022comprehensive}, and forensic analysis~\cite{hadid2012can}, where capturing and tracking individuals without direct interaction is often a priority. Despite its potential, several challenges persist that hinder its effectiveness in real-world environments, including variations in viewpoint, occlusion, and environmental noise.

Traditional approaches to gait recognition typically rely on 2D representations such as silhouettes and skeletons. These methods have become popular due to their simplicity and computational efficiency. 2D representations effectively capture body shape features and joint dynamics, which are essential for recognizing gait. However, these methods face significant limitations when applied in complex, uncontrolled settings. For example, changes in viewpoint or occlusion of key body parts can severely impact the quality of the extracted features, leading to lower recognition accuracy~\cite{chai2022lagrange, 10744527, Gupta_2024_WACV, 10.1007/978-3-031-72658-3_22}. Environmental noise, such as background clutter or lighting variations, further compounds these challenges, making it difficult for 2D-based methods to maintain consistent performance.

\begin{figure}[t]
  \centering
  % \fbox{\rule{0pt}{3in} \rule{1\linewidth}{0pt}}
  % \fbox{}
  {\includegraphics[width=1.0\linewidth]{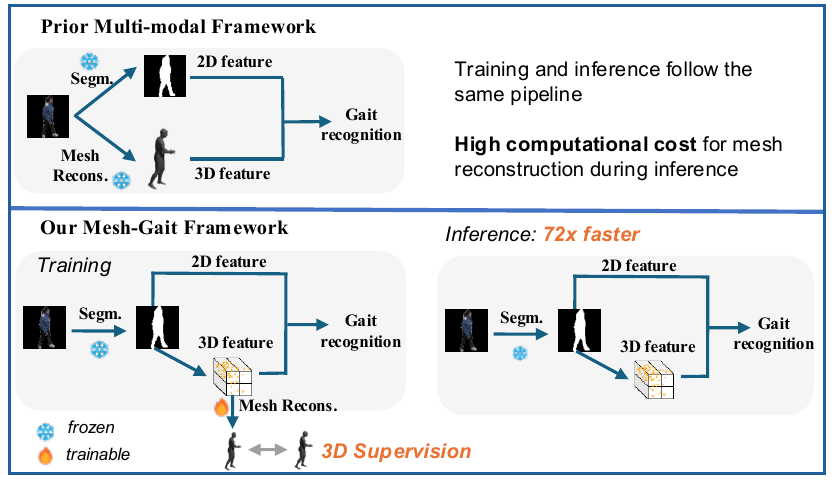}}
   \caption{An example comparison highlighting the differences between the Mesh-Gait framework and traditional multi-modal methods in terms of training, inference, and efficiency. Mesh-Gait is trained from scratch using supervised learning but only requires mask segmentation and mesh reconstruction during training. At inference time, only mask segmentation is required, eliminating the need for mesh reconstruction. Mesh-Gait is more efficient, as it has a lower computational cost since reconstruction is not required during inference.}
   
   \vspace{-4mm}
   \label{fig:teaser}
\end{figure}

To mitigate these limitations, recent research~\cite{zheng2022gait} has proposed 
multi-modal gait recognition frameworks by incorporating 3D body shape information into 2D silhouettes or skeletons. 3D representations offer more detailed structural information, such as depth and body posture, which helps to address the challenges faced by 2D-based methods. multi-modal methods are more robust to viewpoint changes, partial occlusion, and environmental noise, leading to improved accuracy in gait recognition. However, multi-modal approaches come with their own set of challenges. For example, generating accurate 3D gait representations from video data requires significant computational resources and is time-intensive, often making these methods impractical for real-time applications. Current 3D reconstruction techniques typically rely on multi-view cameras or depth sensors, which are not always available or feasible in real-world scenarios. Furthermore, reconstructing 3D body models from RGB videos remains a complex task, often requiring significantly more processing power than simple 2D silhouette segmentation—potentially exceeding the capabilities of typical real-time systems. In addition, extracting dynamic features from 3D representations, such as point clouds or meshes, is challenging, and point-based features are difficult to effectively fuse with silhouette-based features to achieve optimal performance.

To overcome these challenges, we introduce {\bf Mesh-Gait}, a novel end-to-end multi-modal gait recognition framework that directly reconstructs 3D representations from 2D silhouettes. As the example shown in~\Cref{fig:teaser}, unlike conventional methods that rely on high computational cost for mesh reconstruction from RGB videos during inference, Mesh-Gait offers an efficient solution by generating 3D representations directly from 2D silhouette inputs. This approach eliminates the extra need for mesh reconstruction from RGB videos during inference, making it more accessible and practical for real-world applications. 

Furthermore, unlike other methods that directly learn point-based features from 3D skeletons or meshes, Mesh-Gait reconstructs {\bf 3D heatmaps as an intermediate representation} and extracts complementary features from these heatmaps to fuse with silhouette-based features. To ensure the accuracy of the 3D heatmaps, they are also employed for the reconstruction of 3D joints, virtual markers, and 3D meshes. During training, the 3D heatmaps are gradually refined under supervised learning by calculating the loss between the reconstructed 3D joints, virtual markers, and meshes and their corresponding ground truth, ensuring accurate spatial alignment and consistent 3D structure.

Mesh-Gait utilizes a deep learning-based model to extract discriminative features from 2D silhouettes and efficiently reconstruct 3D heatmaps representations. By leveraging both 2D and 3D information, Mesh-Gait enhances gait recognition performance, capturing both the spatial and structural aspects of human gait, which are crucial for accurate identification. The proposed approach is designed to be computationally efficient, reducing the overhead typically associated with 3D reconstruction from RGB video, which makes real-time gait recognition feasible.

We comprehensively evaluate Mesh-Gait on multiple benchmark datasets to assess its effectiveness across different conditions. Our experiments demonstrate that Mesh-Gait not only generates high-quality 3D gait representations from 2D silhouettes but also significantly outperforms state-of-the-art methods in terms of both recognition accuracy and robustness. Specifically, Mesh-Gait demonstrates strong performance in challenging scenarios, such as variations in viewpoint, partial occlusion, and environmental noise, where traditional 2D methods struggle. Moreover, Mesh-Gait achieves a level of computational efficiency that makes real-time gait recognition practical, even in environments with limited computational resources. 

The key contributions of this work are fourfold:  

1. Mesh-Gait introduces a novel multi-modal framework that generates high-quality 3D gait representations directly from 2D silhouettes, eliminating the need for complex multi-view or depth-sensing setups and mesh reconstruction from RGB videos during inference. 

2. Mesh-Gait reconstructs 3D heatmaps as an intermediate representation and extracts complementary features from these heatmaps to fuse with silhouette-based features. 

3. The intermediate 3D heatmaps are gradually refined during supervised learning by using them to reconstruct 3D joints, virtual markers, and meshes and calculating the loss between these reconstructions and their corresponding ground truth, thereby ensuring precise spatial alignment and consistent 3D structure.

4. Through extensive evaluation, Mesh-Gait outperforms existing state-of-the-art methods in recognition accuracy, robustness, and computational efficiency.

\section{Related Works}
\label{sec:formatting}

\begin{figure*}[t]
  \centering
  % \fbox{\rule{0pt}{3in} \rule{1\linewidth}{0pt}}
  % \fbox{}
  {\includegraphics[width=1.0\linewidth]{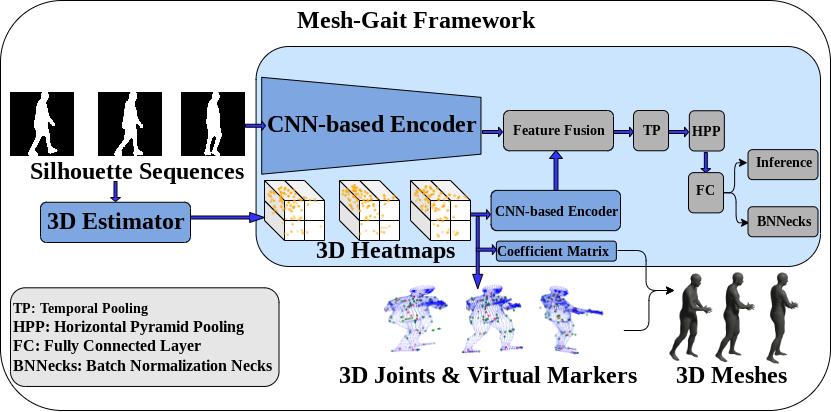}}
   \caption{{\bf The main architecture of Mesh-Gait} consists of two parallel branches that process silhouette sequences extracted from RGB videos using an image segmentation model. The 2D feature branch employs a convolutional backbone to extract gait features from 2D silhouettes. In parallel, the 3D feature branch reconstructs 3D heatmaps as an intermediate representation from the silhouette sequences using a 3D estimator, which is trained from scratch. To progressively refine the 3D heatmaps during training, they are used for reconstructing 3D joints, virtual markers, and meshes in a supervised manner. In addition, the reconstructed 3D heatmaps are also used for 3D feature extraction. Features from both branches are then fused and mapped for gait recognition. The model is trained in a supervised manner using a combination of triplet loss, cross-entropy loss, L1 loss, and L2 loss.}
   
   \vspace{-4mm}
   \label{fig:vm_pipline}
\end{figure*}

Several notable advancements have been made in silhouette-based and skeleton-based gait recognition~\cite{fan2023opengait,wang2003silhouette,guo2023multi,makihara2011gait,wang2024hypergait,li2020jointsgait,huang2023condition,teepe2021gaitgraph,teepe2022towards,addabbo2021temporal,liao2020model}. These approaches commonly leverage deep learning architectures such as Convolutional Neural Networks to extract gait features. Specifically, GaitSet~\cite{chao2021gaitset} and GaitPart~\cite{fan2020gaitpart} introduced CNN-based approaches to extract spatial-temporal gait features. GaitGL~\cite{lin2021gait} introduced a combination of global and local feature learning to improve recognition. More recently, a flexible gait recognition codebase named OpenGait~\cite{fan2023opengait} has been proposed. DeepGaitV2~\cite{fan2023exploring} uses CNNs, Transformers, and multi-scale feature extraction for improved gait representation. Despite their effectiveness, these methods are influenced by factors like occlusions, viewpoint variation, and environmental noise. Some multimodal gait recognition methods combining 2D-based and 3D-based representations information are proposed to mitigate the aforementioned issues such as LiCamGait~\cite{han2022licamgait}, LidarGait~\cite{shen2022lidar}, GaitPoint~\cite{chen2022gaitpoint} and SMPLGait~\cite{li2022multi}. These hybrid approaches boost recognition accuracy and strengthen robustness against real-world variations. However, it also introduces challenges in terms of computational complexity and resource demands, which can hinder large-scale deployment. To address these challenges, we introduce a novel multimodal gait recognition framework, Mesh-Gait.

\section{Methodology}

\subsection{Overview}
The main architecture of Mesh-Gait is illustrated in~\Cref{fig:vm_pipline}. Mesh-Gait consists of a Dual-Branch backbone, comprising a 2D feature branch and a 3D feature branch. The 2D feature branch uses a convolutional-based encoder to extract 2D gait features from 2D silhouettes. The 3D feature branch, on the other hand, performs 3D mesh reconstruction from the 2D silhouettes and then extracts 3D gait features from reconstructed 3D models. The features obtained from both branches are then fused and mapped for gait recognition. To train the model, we employ a combination of triplet loss, cross-entropy loss, L1 loss, and L2 loss.

\subsection{Network Structure}

Let the input silhouette sequences be represented as

\( I \in \mathbb{R}^{B \times C \times T \times H \times W} \), where: 
\begin{itemize}
    \item \( B \) is the batch size; \( C \) is the number of input channels,
    \item \( T \) is the time length of the silhouette sequences,
    \item \( H \) and \( W \) are the height and width of each silhouette.
\end{itemize}

{\bf The 2D feature branch} applies a feature extraction module consisting of a series of convolutional layers and residual blocks to extract features. Let \( F_0 \) be the initial feature map after the feature extraction module:
\[
F_0 = \text{ResidualBlock}(I) \in \mathbb{R}^{B \times T \times C^{p} \times H' \times W' }
\]
\begin{itemize}
    \item \( C^{p} \) is the number of output channels,
    \item \( T \) is the time length of the silhouette sequences,
    \item \( H' \) and \( W' \) are the down-sample size of the height and width of each silhouette.
\end{itemize}
{\bf The 3D feature branch} applies a 3D estimator to estimate 3D heatmaps corresponding to their 2D silhouettes for capturing spatiotemporal relationships. HRNet~\cite{sun2019deep} is employed as the 3D estimator due to its high-resolution representations and strong performance in keypoint detection tasks.  
Let the heatmaps 
\[ H \in \mathbb{R}^{B \times T \times C'' \times D'' \times H' \times W'} \] represent the output of the heatmap module, where:
\begin{itemize}
    \item \( D' \), \( H' \) and \( W' \) are the spatial dimensions of the heatmap; \( C' \) is the number of heatmap channels.
\end{itemize}

The generated 3D heatmaps are simultaneously used for 3D gait feature extraction and 3D representation reconstruction, ensuring that reconstructed 3D heatmaps become increasingly accurate during supervised training.
(1) 3D representation reconstruction:
Each generated 3D heatmap encodes the per-voxel likelihood of each joint and virtual marker~\cite{ma20233d}. The concept of virtual markers was initially introduced as an intermediate representation for 3D human mesh estimation. Inspired from physical markers placed on the body for mesh reconstruction, virtual markers simulate these markers by identifying a set of key vertices on the 3D mesh. Unlike mesh vertices, the number of virtual markers is much smaller, providing a more efficient representation while retaining the essential information required for accurate body pose and movement reconstruction. The 3D positions of the joints and virtual markers are represented as:
\[
(J,V) = \sum^{D'}_{d=1}\sum^{H'}_{h=1}\sum^{W'}_{w=1} (d,h,w) \cdot H(d,h,w)  
\]
where 3D joint and virtual marker 
\[ (J , V) \in \mathbb{R}^{B \times T \times C' \times (J+V) \times 3} \].

Then, the 3D meshes 
\[ Meshes \in \mathbb{R}^{B \times T \times C' \times M \times 3} \] can be computed based on the joints and virtual markers multiplied with a learnable coefficient matrix C :
\[
Meshes = (J , V) \times C  
\]
\begin{itemize}
    \item \( J \), \( V \) and \( M \) are the vertex number of joint, virtual marker and mesh; 3 are the spatial position \(x,y,z\),
   
\end{itemize}

To provide more details of the network structure, the pseudocode of 3D representation reconstruction from silhouettes is also provided in Algorithm~\ref{alg:cond_gen}.

(2) 3D gait feature extraction:

Instead of extracting point-based features from 3D joints, virtual markers, and meshes, we extract 3D gait features from 3D heatmaps. The generated 3D heatmaps \(H\) also passed through a heatmap processing module consisting of 3D convolutions, batch normalization, and ReLU activations:
\[
H_{\text{processed}} = \text{Conv3d}(H) \quad \in \mathbb{R}^{B \times T \times C^{r} \times H' \times W' }
\]
To further extract spatiotemporal features from heatmaps, we use a similar feature extraction module as 
 the 2D feature branch:
\[
F_1 = \text{ResidualBlock}(H_{\text{processed}}) \quad \in \mathbb{R}^{B \times T \times C^{q} \times H' \times W' }
\]

\begin{itemize}
    \item \( C^{r} \) and \( C^{q} \) are the number of output channel of \( H_{\text{processed}} \) and \( F_1 \).
\end{itemize}

\begin{figure}[]
\centering
\begin{minipage}{0.45\textwidth}
\begin{algorithm}[H]
    \caption{3D Representation Reconstruction from Silhouettes}
    \begin{algorithmic}[1]
    \STATE \textbf{Input:} Silhouette sequence $S \in \mathbb{R}^{B \times T \times 64 \times 44}$ 
    \STATE \textbf{Output:} 3D mesh coordinates $M \in \mathbb{R}^{B \times T \times 6890 \times 3}$ 
    
    \STATE \textbf{Step 1: Heatmap Generation}
    \STATE Get 3D heatmaps from silhouettes: $H \gets \text{HRNet}(S)$
    \STATE Reshape $H$ to shape $(BT, J+V, D, H, W)$
    \STATE \textbf{Step 2: Joint \& Marker Localization}
    \STATE $H_x \gets \sum_{d,h} H$ \hfill  {\color{gray}{\# Sum over depth and height }}
    \STATE $H_y \gets \sum_{d,w} H$ \hfill {\color{gray}{\# Sum over depth and width }}
    \STATE $H_z \gets \sum_{h,w} H$ \hfill {\color{gray}{\# Sum over height and width}}
    \STATE $x \gets \sum_w H_x$, $y \gets \sum_h H_y$, $z \gets \sum_d H_z$ 
    
    {\color{gray}{\# $x,y,z$→ $(BT, J+V, 1)$ each}}
    \STATE $(J,V) \gets \text{concat}(x, y, z)$  
    
    {\color{gray}{\# Joint \& Marker coords → $(BT, J+V, 3)$}}
    
    \STATE \textbf{Step 3: Mesh Regression}
    \STATE Compute index $I \gets z \cdot H \cdot W + y \cdot W + x$ \hfill 
    
    {\color{gray}{\# → $(BT, J+V, 1)$}}
    \STATE $H_{\text{flat}} \gets \text{reshape}(H, (BT, J+V, D \cdot H \cdot W))$
    \STATE $\alpha \gets \text{gather}(H_{\text{flat}}, I)$ \hfill 
    
    {\color{gray}{\# Confidence scores → $(BT, J+V)$}}
    \STATE Initialize adaptive regressor: $\mathcal{A} \gets \text{Linear}(J+V, 6890 \cdot (J+V))$
    \STATE $Coef \gets \text{reshape}(\mathcal{A}(\alpha), (BT, 6890, J+V))$
    \STATE $M \gets Coef \cdot (J,V)$ 
    
     \hfill {\color{gray}{\# Final mesh coordinates → $(BT, 6890, 3)$}}
    \end{algorithmic}
    \label{alg:cond_gen}
\end{algorithm}
\end{minipage}

\end{figure}

\subsection{Multi-modal feature fusion}
We use the concatenation operation to fuse features from 2D feature branch and 3D feature branch:
\[
F_2 = Cat(F_0, F_1)  \in \mathbb{R}^{B \times T \times C^{(p+q)} \times H' \times W' } \
\]
and processed through temporal pooling \(TP\) 
\[
F_3 = TP(F_2)  \in \mathbb{R}^{B \times C^{(p+q)} \times H' \times W' } \
\]
and horizontal pyramid pooling \(HPP\) to aggregate spatiotemporal information.
\[
F_4 = HPP(F_3)  \in \mathbb{R}^{B \times C^{(p+q)} \times P } \
\]
\begin{itemize}
    \item \( P \) is the number of mapped features.
\end{itemize}

\subsection{Loss Function}
To train the Mesh-Gait model, we employ a combination of triplet loss, cross-entropy loss, L1 loss, and L2 loss. The triplet loss and cross-entropy loss are used for gait recognition while the L1 loss and L2 loss are used for 3D joints, virtual markers and meshes reconstruction to help refine 3D heatmaps during training.

\textbf{Triplet Loss:} the embeddings \( E \in \mathbb{R}^{B \times C \times P} \) are calculated from the feature map:
\[
E = \text{FC}(F_H)
\]
The triplet loss can be expressed as:
\[
L_{\text{triplet}} = \sum_{i} \left[ \max(\| E_i - E_{\text{anchor}} \|_2^2 - \| E_i - E_{\text{positive}} \|_2^2 + \alpha), 0 \right]
\]
Where \( \alpha \) is a margin, \( E_{\text{anchor}} \) is the embedding of the anchor sample, and \( E_{\text{positive}} \) is the embedding of the positive sample.

\textbf{Cross-entropy Loss:} For classification tasks, we calculate the \( {\text{logits}} \) by using \(BNNecks\):
\[
{\text{logits}} = \text{BNNecks}(E)
\]
The cross-entropy loss is:
\[
L_{\text{softmax}} = - \sum_{i} y_i \log({\text{logits}_i})
\]
Where \( y_i \) is the ground truth label for the \( i \)-th sample.

\textbf{Joint and Mesh Loss:} For 3D joints and virtual markers, the MSE loss is:
\[
L_{\text{mse\_joint}} = \frac{1}{B} \sum_{i} \| J_i - J_{\text{gt}} \|_2^2
\]
For mesh loss, the L1 loss is :
\[
L_{\text{mesh}} = \frac{1}{B} \sum_{i} \| M_i - M_{\text{gt}} \|_1
\]

\textbf{Total Loss:}
The total loss is a weighted sum of all individual losses:
\[
L_{\text{total}} = \lambda_1 L_{\text{triplet}} + \lambda_2 L_{\text{softmax}} + \lambda_3 L_{\text{mse\_joint}} + \lambda_4 L_{\text{mesh}}
\]
Where \( \lambda_1, \lambda_2, \lambda_3, \lambda_4 \) are the loss weights.

\begin{table*}[ht]
\centering
\small
\renewcommand{\arraystretch}{1.3}
\setlength{\tabcolsep}{0.5cm}
\begin{tabular}{c|c|cccc}
\hline
\multicolumn{1}{c|}{{Gait Representations}} & Methods  & R-1  & R-5 & mAP  & mINP  \\\hline
\multirow{2}{*}{Skeletons} 
& PoseGait~\cite{liao2020model} & 0.2 & 1.1 & 0.5 & 0.3 \\
& GaitGraph~\cite{teepe2021gaitgraph} & 6.3 & 16.2 & 5.2 & 2.4 \\\hline
\multirow{8}{*}{Silhouettes} 
& GEINet~\cite{shiraga2016geinet} & 5.4 & 14.2 & 5.1 & 3.1 \\
& GaitSet~\cite{chao2021gaitset} & 36.7 & 58.3 & 30.0 & 17.3 \\
& GaitPart~\cite{fan2020gaitpart} & 28.2 & 47.6 & 21.6 & 12.4 \\
& GLN~\cite{hou2020gait} & 31.4 & 52.9 & 24.7 & 13.6 \\
& GaitGL~\cite{lin2021gait} & 29.7 & 48.5 & 22.3 & 13.3 \\
& CSTL~\cite{huang2021context} & 11.7 & 19.2 & 5.6 & 2.6 \\
& SMPLGait w/o 3D~\cite{zheng2022gait} & 42.9 & 63.9 & 35.2 & 20.8 \\
& GaitBase~\cite{fan2023opengait} & 64.6 & - & - & - \\
& DeepGaitV2-P3D~\cite{fan2023exploring} &\cellcolor{second} 74.4 & \cellcolor{best}\bf 88.0 &\cellcolor{second} 65.8 & - \\\hline

\multirow{2}{*}{Multi-Modalities} 

& SMPLGait~\cite{zheng2022gait} & 46.3 & 64.5 & 37.2 & \cellcolor{second}22.2 \\

& \bf Mesh-Gait & \cellcolor{best}\bf 75.0 & \cellcolor{best}\bf 88.0 & \cellcolor{best}\bf 66.1 &\cellcolor{best} \bf 39.5 \\\hline
\end{tabular}
\caption{We present the gait recognition results on the Gait3D dataset, where we compare our proposed Mesh-Gait framework with twelve state-of-the-art methods. The experiments were conducted with an input size of (64 × 44). Mesh-Gait outperforms all the compared methods, showcasing superior performance. This result highlights the effectiveness of our multi-modal framework in reconstructing and leveraging complementary 3D representations alongside traditional silhouette-based features}
\label{gait3d}
% \vspace{-4mm}
\end{table*}

\begin{table*}
\centering
\setlength{\tabcolsep}{0.2cm}
\small
\renewcommand{\arraystretch}{1.3}
\begin{tabular}{c|ccccccccccccccc}\hline
\multicolumn{1}{c|}{}  &\multicolumn{14}{c}{Probe View} &\multicolumn{1}{c}{}\\\hline
Methods            & $0^{\circ}$ & $15^{\circ}$ & $30^{\circ}$ & $45^{\circ}$ & $60^{\circ}$& $75^{\circ}$& $90^{\circ}$& $180^{\circ}$& $195^{\circ}$& $210^{\circ}$& $225^{\circ}$& $240^{\circ}$& $255^{\circ}$& $270^{\circ}$ &Mean\\\hline

SMPLGait~\cite{zheng2022gait} &52.1&67.6&74.7&77.0&71.4&73.1&69.6&53.9&67.4&73.5&75.2&69.9&71.1&66.8&68.8 \\
SMPLw/o3D~\cite{zheng2022gait} &56.2&73.9&81.0&82.2&75.7&78.3&75.4&60.1&74.0&80.2&81.3&74.9&76.9&73.3&74.5 \\
GaitSet~\cite{chao2021gaitset}  &65.8&80.1&85.0&85.5&81.0&82.8&81.2&69.3&79.8&83.9&84.7&79.8&81.2&78.8&79.9\\
GaitPart~\cite{fan2020gaitpart} &68.6&81.9&86.3&86.7&83.1&84.3&82.9&72.3&81.4&85.1&85.6&82.2&82.8&80.8&81.7\\
GaitGL~\cite{lin2021gait} &71.2&82.4&86.1&86.6&83.3&84.8&83.6&75.4&81.8&85.0&85.7&82.1&83.2&81.5&82.3\\

GaitBase~\cite{fan2023opengait}        &75.1&85.6&88.6&89.0&85.9&86.9&85.7&78.1&85.4&87.8&88.4&85.2&85.8&84.2&85.1 \\

DeepGaitV2~\cite{fan2023exploring} &\cellcolor{second}83.2&\cellcolor{second} 88.9& \cellcolor{best}\bf90.2& \cellcolor{second}90.5&\cellcolor{second}89.3&\cellcolor{second}89.4&\cellcolor{second}89.0&\cellcolor{second}85.9&\cellcolor{second}88.3&\cellcolor{second}89.3&\cellcolor{second}89.8&\cellcolor{second}88.5&\cellcolor{second}88.3&\cellcolor{second}87.7&\cellcolor{second}88.5\\
\bf Mesh-Gait        &\cellcolor{best}\bf83.7&\cellcolor{best}\bf89.1&\cellcolor{best}\bf90.2&\cellcolor{best}\bf90.6&\cellcolor{best}\bf89.7&\cellcolor{best}\bf89.5&\cellcolor{best}\bf89.2&\cellcolor{best}\bf86.5&\cellcolor{best}\bf88.6&\cellcolor{best}\bf89.5&\cellcolor{best}\bf89.9&\cellcolor{best}\bf88.7&\cellcolor{best}\bf88.4&\cellcolor{best}\bf87.9&\cellcolor{best}\bf88.7\\

\hline

\end{tabular}

\caption{Gait recognition results on the OUMVLP-Mesh~\cite{li2022multi} dataset. We compare Mesh-Gait with seven state-of-the-art methods. The input size of silhouette sequences is $64 \times 44$. To evaluate the model's performance, we compute the Rank-1 accuracy under 14 probe views, excluding identical-view cases. Mesh-Gait consistently demonstrates superior performance, achieving state-of-the-art results across all probe views. This highlights its effectiveness in handling various viewing angles and showcases its robustness in gait recognition.}
\label{oumvlp}
% \vspace{-2mm}
\end{table*}

\begin{table*}
\small
\centering
\renewcommand{\arraystretch}{1.3}
\scalebox{1.0}{
\begin{tabular}{c|cccc} 
% \toprule
\hline

\multirow{2}{*}{Methods}    & \multicolumn{4}{c}{Recognition Metrics(\%)}\\
            & R-1 & R-5 & mAP & mINP  \\\hline
GaitPart~\cite{fan2020gaitpart}           & 28.2     & 47.6     & 21.6    & 12.4 \\
GaitPart~\cite{fan2020gaitpart} w/ Mesh-Gait    &\bf31.5	&\bf50.6	&\bf24.1	&\bf13.2 \\\hline
GaitSet~\cite{chao2021gaitset}                & 36.7     & 58.3     & 30.0    &17.3  \\
GaitSet~\cite{chao2021gaitset} w/ Mesh-Gait                 &\bf43.0	&\bf62.2	&\bf71.9	&\bf19.19 \\\hline
SMPLGait w/o 3D~\cite{zheng2022gait}    &42.9 & 63.9 & 35.2 & 20.8\\
SMPLGait~\cite{zheng2022gait}            & 46.3     & 64.5     & 37.2    & \bf 22.2 \\
SMPLGait~\cite{zheng2022gait} w/ Mesh-Gait    & \bf 51.4     & \bf 69.4       & \bf 41.3    & \bf22.2 \\\hline
GaitBase~\cite{fan2023opengait}              & 64.6     & -     & -    & - \\
GaitBase~\cite{fan2023opengait}  w/ Mesh-Gait            &\bf64.9     &\bf79.5       &\bf54.4    &\bf31.1 \\\hline
DeepGaitV2~\cite{fan2023exploring} & 74.4     & \bf88.0     & 65.8    & - \\
\bf DeepGaitV2~\cite{fan2023exploring} w/  Mesh-Gait  & \bf75.0     &\bf 88.0     & \bf 66.1    & \bf 39.5 \\
\hline
% \bottomrule

\end{tabular}}
\caption{We conducted a series of experiments to switch five different silhouette
backbones used in state-of-the-art methods. The results clearly demonstrate that the Mesh-Gait framework consistently delivers better performance, even when switching between different silhouette backbones. This highlights the robustness and flexibility of the Mesh-Gait approach, showcasing its ability to maintain high recognition accuracy across varying backbone configurations.}
\label{backbones}
\end{table*}

\section{Experiments}

We conducted a comprehensive series of experimental evaluations to thoroughly assess the performance and capabilities of our proposed Mesh-Gait methodology. The first phase of our experimentation focused on extensive benchmarking, where we systematically compared Mesh-Gait against current state-of-the-art approaches across multiple representation modalities. These comparative analyses were performed using two significant and widely-recognized public 3D gait datasets: ${\bf Gait3D}$~\cite{zheng2022gait} and ${\bf OUMVLP-Mesh}$~\cite{li2022multi}, which provided robust testing environments for our evaluations. Following the benchmarking phase, we proceeded with detailed ablation studies designed to thoroughly investigate two key aspects: first, the contribution and effectiveness of the multi-modal branch within our architecture, and second, the framework's inherent flexibility and adaptability when incorporating various types of silhouette encoder backbones. This systematic experimental approach allowed us to comprehensively validate both the overall performance and the individual components of our proposed method.

\subsection{Experimental Setup}

\subsubsection{Gait Datasets}

Gait3D~\cite{zheng2022gait} is a groundbreaking large-scale dataset that is collected for real-world gait recognition. It contains both 2D silhouettes and 3D human meshes collected from 4,000 individuals, comprising 25,309 sequences in total. The 3D data are estimated from video frames. In our research, we focused on a comprehensive subset involving all 4,000 subjects, specifically working with their 2D silhouettes and corresponding 3D body vertex data.

OUMVLP-Mesh~\cite{li2022multi} is derived from the OU-ISIR Gait Database and serves as a large-scale multi-view dataset containing human mesh data. The dataset covers 10,307 subjects and captures them from 14 different viewing angles. These angles span from 0° to 90° and 180° to 270°, with measurements taken at 15° intervals. While OUMVLP-Mesh builds upon the original OUMVLP dataset, it's worth noting that not all sequences include 3D information. For our analysis, we specifically selected frames that contained both silhouette and 3D data, which resulted in a smaller total frame count compared to the original OUMVLP dataset.

\subsubsection{Implementation Details}

We conduct a comparative analysis of Mesh-Gait against state-of-the-art methods, categorized as follows: Skeleton-based methods encompass PoseGait~\cite{liao2020model} and GaitGraph~\cite{teepe2021gaitgraph}; Silhouette-based methods include GEINet~\cite{shiraga2016geinet}, GaitSet~\cite{chao2021gaitset}, GaitPart~\cite{fan2020gaitpart}, GLN~\cite{hou2020gait}, GaitGL~\cite{lin2021gait}, CSTL~\cite{huang2021context}, GaitBase~\cite{fan2023opengait} and DeepGaitV2~\cite{fan2023exploring}; and the Multi-modality representation method is SMPLGait~\cite{zheng2022gait}.

The input resolution of the 2D silhouette sequences is standardized to 64 × 44. In the 2D feature branch, we utilize the DeepGaitV2~\cite{fan2023exploring} model as the backbone for feature extraction. In the 3D branch, the spatial resolution of the generated 3D heatmaps is set to $16\times16\times16$, The number of reconstructed 3D joint keypoints is 24; the number of reconstructed 3D virtual marker vertices is 64; and the number of reconstructed 3D mesh vertices is 6,890. HRNet-W48~\cite{sun2019deep} is used as the 3D estimator in our experiment. The residual block used for extracting features from the 3D heatmaps in the 3D branch is also based on the DeepGaitV2~\cite{fan2023exploring} backbone. During training, the batch size for the Gait3D dataset is set to $32 \times 4$, where 32 represents the number of subjects, and 4 denotes the number of training sequences per subject. The batch size for the OUMVLP-Mesh dataset during training is set to $32 \times 8$. To facilitate a fair comparison with state-of-the-art methods under the same configuration settings, we utilize the ${OpenGait}$~\cite{fan2023opengait} codebase. Further details regarding training configurations and parameters are included in the Supplementary Material.

\begin{table}[]
\centering
\small
\renewcommand{\arraystretch}{1.3}

\begin{tabular}{c|cccc}
\hline
{{Fusion Strategy}} & {{R-1}} & {{R-5}} & {{mAP}} & {{mINP}} \\
\hline
{Attention} & {73.9} & {86.9} & {65.1} & {38.4} \\
{Add} & {74.1} & {86.0} & {63.6} & {37.2} \\
{Concatenate} & {\bf 75.0} & {\bf 88.0} & {\bf 66.1} & {\bf 39.5} \\
\hline
\end{tabular}

\caption{Ablation study on the effect of different fusion strategies in the Mesh-Gait framework using the Gait3D dataset. The results show that concatenation-based fusion of silhouette and 3D representation features achieves the best performance across all evaluation metrics.}
\vspace{-6mm}
\label{fusion}
\end{table}
\subsubsection{Evaluation Protocol}

In our evaluation process, we followed a standardized experimental setup aligned with best practices in the field. The dataset was divided into training and testing subsets. For the Gait3D dataset, a total of 4,000 subjects were used, with 3,000 allocated for training and 1,000 for testing. In the OUMVLP-Mesh dataset, 5,153 subjects were used for training, while 5,154 subjects were used for testing.

For testing, we employed a probe-gallery evaluation framework. In this setup, one sequence was randomly selected as the probe for each subject, while all remaining sequences from the same subject were assigned to the gallery.

Performance was measured using multiple evaluation metrics. For the Gait3D dataset, the model's effectiveness is assessed by using average Rank-1 and Rank-5 recognition accuracy, mean average precision (mAP), and mean inverse negative penalty (mINP). In the case of the OUMVLP-Mesh dataset, Rank-1 accuracy was computed across 14 different viewing angles, where the probe and gallery views were identical being excluded from evaluation.

\section{Main Results}

\subsection{Comparison with Other State-of-the-Arts Methods}

We present a comprehensive comparison of Mesh-Gait with state-of-the-art methods on the Gait3D and OUMVLP-Mesh datasets in~\Cref{gait3d} and~\Cref{oumvlp}, respectively.

On the Gait3D dataset, Mesh-Gait outperforms existing methods, achieving a Rank-1 accuracy of 75.0\%, the highest among all evaluated approaches. This result highlights the effectiveness of our multi-modal framework in reconstructing and leveraging complementary 3D representations alongside traditional silhouette-based features. Notably, Mesh-Gait is trained entirely from scratch, demonstrating its ability to learn robust gait representations without relying on pre-trained models. Compared to DeepGaitV2~\cite{fan2023exploring}, Mesh-Gait exhibits superior feature extraction and fusion capabilities. Unlike other multi-modal methods that require an additional human mesh reconstruction model from RGB videos to generate 3D information, our approach directly performs 3D mesh reconstruction and gait recognition simultaneously. This end-to-end design enhances efficiency and accuracy, making Mesh-Gait a strong candidate for real-world applications.

On the OUMVLP-Mesh dataset, as detailed in~\Cref{oumvlp}, Mesh-Gait sets a new benchmark by achieving state-of-the-art performance across all probe views, with a mean Rank-1 accuracy of 88.7\%. The model’s dual-branch architecture proves highly effective, as it learns viewpoint-invariant features from the reconstructed 3D representations, which provide complementary structural and motion information to silhouette-based features. This integration enhances recognition robustness across diverse viewing angles, allowing Mesh-Gait to outperform existing methods in capturing fine-grained gait dynamics.
As mentioned in the experimental setup, OUMVLP-Mesh is built upon the original OUMVLP dataset. However, not all sequences in the OUMVLP dataset contain corresponding 3D information. As a result, OUMVLP-Mesh has a reduced number of frames compared to the OUMVLP dataset. This discrepancy in frame count impacts gait recognition performance, leading to differences in results between the two datasets.

\begin{table}[]
 \centering
 \small
 \renewcommand{\arraystretch}{1.3}
\begin{tabular}{ccc|cccc}
\hline
Heatmap&joint&mesh   &R-1      & R-5   & mAP    & mINP        \\ \hline	

$\checkmark$ &$\checkmark$ &$\times$ & 73.4 & 87.0 & 64.8 & 38.2	\\
$\checkmark$ &$\checkmark$ &$\checkmark$ & \bf 75.0 & \bf88.0 & \bf66.1 & \bf 39.5 \\\hline

\end{tabular}

\caption{Ablation study on the effect of different 3D reconstruction representations in the Mesh-Gait framework. The results show that incorporating 3D meshes alongside heatmaps and 3D joints \& virtual markers leads to better performance, highlighting the importance of full 3D reconstruction for improved gait recognition.}

\label{selection}
\end{table}

\begin{table}[]
\renewcommand{\arraystretch}{1.3}
\small
\centering
\begin{tabular}{c|cccc}
\hline
Feautre Dimension  & R-1  & R-5 & mAP  & mINP  \\\hline

4 & 74.9 & 87.2 & \bf66.3 & \bf39.7 \\
8 & \bf75.0 &\bf88.0 & 66.1 &39.5 \\
16 & 73.8 &87.1 & 65.5 & 38.7 \\\hline
\end{tabular}
\caption{Ablation study on the effect of 3D heatmap feature channel dimension in the Mesh-Gait framework. The results highlight the importance of selecting an optimal feature dimension.}
\label{Feautre_Dimension}
\vspace{-8mm}
\end{table}

\subsection{Ablation Study}
\subsubsection{Analysis of Mesh-Gait Robustness} 

In this section, we investigate the robustness and adaptability of Mesh-Gait across various feature extraction architectures and whether its multi-modal design consistently enhances recognition performance. To achieve this, we integrate five representative silhouette-based feature extraction networks widely used in state-of-the-art gait recognition methods: GaitSet~\cite{chao2021gaitset}, GaitPart~\cite{fan2020gaitpart}, SMPLGait~\cite{zheng2022gait}, Gaitbase~\cite{fan2023opengait} and DeepGaitV2~\cite{fan2023exploring}. All experiments are conducted on the Gait3D dataset to ensure a consistent evaluation framework. The results of these experiments are summarized in~\Cref{backbones}. The integration of Mesh-Gait consistently enhances the performance of various gait recognition methods across all evaluation metrics, demonstrating its effectiveness in improving gait recognition. This suggests that our framework can offer stable enhanced complementary information to different silhouette features, thereby improving recognition performance. Mesh-Gait demonstrates strong adaptability for real-world applications.

\begin{figure}[t]
  \centering
  % \fbox{\rule{0pt}{3in} \rule{1\linewidth}{0pt}}
  % \fbox{}
  {\includegraphics[width=0.5\textwidth]{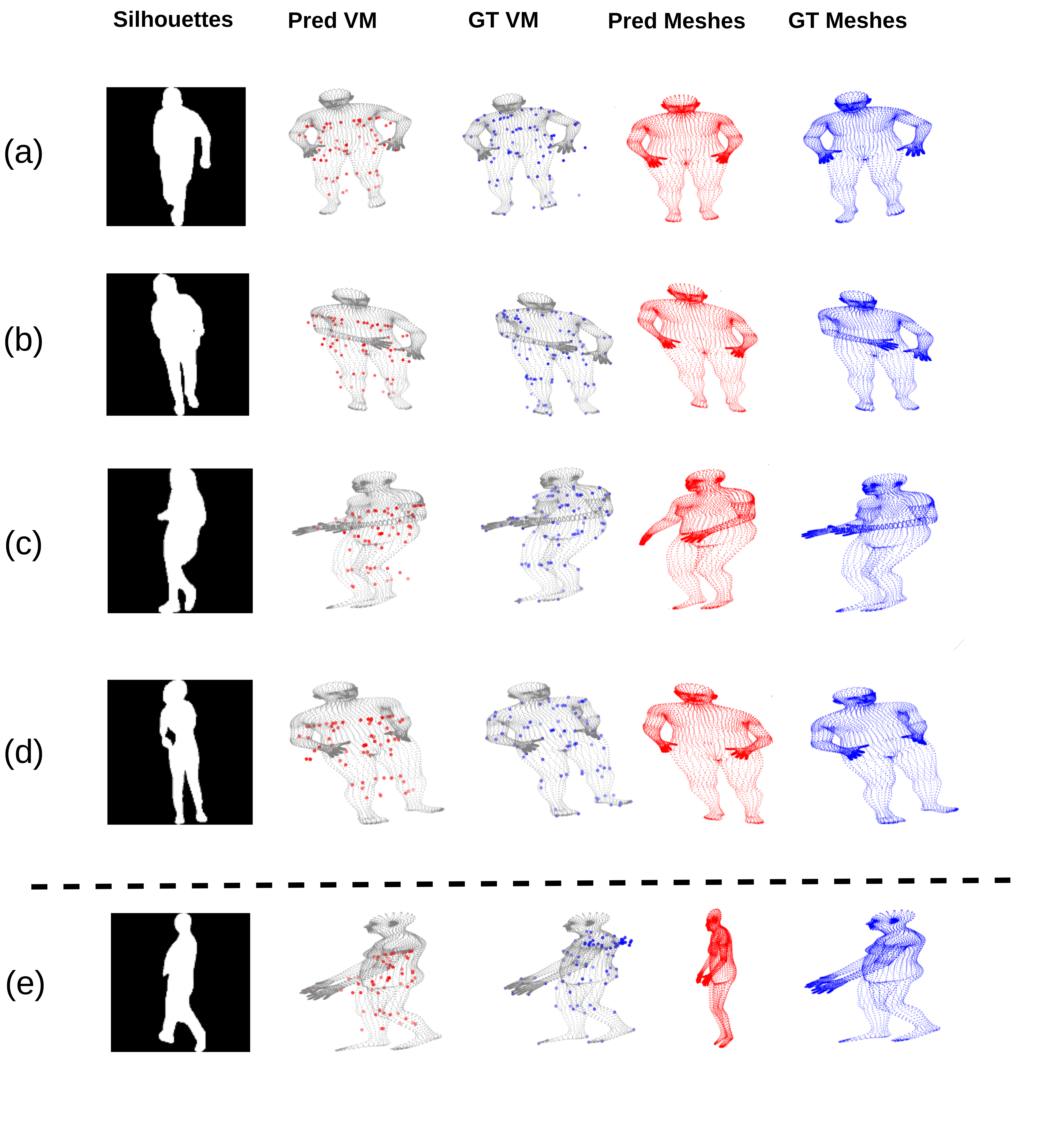}}
   \caption{Visualization of reconstructed 3D representations during testing. Each row represents the reconstruction results on a different subject. The predicted virtual markers and ground truth virtual markers are shown on the surface of the ground truth meshes. The predicted meshes reconstructed from predicted virtual markers are shown in the fourth column. The ground truth of Meshes is shown in the fifth column.}
   \label{fig:plot_results}
   % \vspace{-4mm}
\end{figure}

\subsubsection{Effect of Fusion Strategy}
We also explore the impact of different fusion strategies for integrating multi-modal features within the Mesh-Gait framework. To this end, we conduct experiments using three fusion methods: attention-based fusion, addition-based fusion, and concatenation-based fusion. As presented in \Cref{fusion}, the results indicate that concatenating silhouette features with 3D representation features is both a simple and highly effective strategy. This approach yields the best performance across all evaluation metrics, demonstrating its strength in combining complementary information from both 2D and 3D modalities.

\subsubsection{Effect of Different 3D Representations}
To further analyze the impact of different 3D representations in the Mesh-Gait framework, we conduct an ablation study by comparing two configurations: Full 3D Reconstruction: Incorporating heatmaps, 3D joints \& virtual markers, and 3D meshes. Partial 3D Reconstruction: Utilizing only heatmaps and 3D joints \& virtual markers without 3D meshes. The results, summarized in \Cref{selection}, indicate that excluding the 3D mesh leads to a noticeable drop in performance. This suggests that reconstructing the 3D mesh enhances the effectiveness of 3D heatmaps by providing richer and more structured complementary information. Consequently, leveraging the full 3D representation improves the overall gait recognition performance.

\subsubsection{Effect of Hyperparameters}
We further investigate the impact of hyperparameters in the Mesh-Gait framework, starting with the effect of the 3D heatmap feature channel dimension on performance. We conduct experiments using three different feature dimensions: 4, 8, and 16. The results, summarized in \Cref{Feautre_Dimension}, indicate that a feature dimension of 8 achieves the highest Rank-1 and Rank-5 accuracy, while a dimension of 4 yields the best mAP and mINP scores. However, increasing the feature dimension to 16 results in a slight decline in performance across all metrics. These findings highlight the importance of selecting an optimal feature dimension to balance compactness and expressiveness, ensuring effective gait representation.

\subsubsection{Visualization of Reconstued 3D Representations} 
\label{sec:VisualizationComplementarity}
To demonstrate the accuracy of the reconstructed 3D representations in MeshGait, we present both the reconstructed 3D outputs and the ground truth in~\Cref{fig:plot_results}. The results clearly show that the reconstructed 3D joints, virtual markers, and meshes align closely with the ground truth, indicating high accuracy. A failure case is also highlighted in panel (e), which may be due to ambiguities in the 2D silhouettes. Additionally, we examine how the reconstructed 3D information complements the 2D representations. The results reveal that even in situations where the 2D silhouette contains noise or is incomplete, MeshGait is still able to accurately identify individuals. This suggests that the 3D features in MeshGait are robust to noise and incompleteness, while providing valuable complementary information to the 2D silhouette features.

\subsection{Efficiency Analysis of Mesh-Gait}
Removing mesh reconstruction at inference time reduces computational overhead, resulting in faster processing and lower latency for real-world deployment. To evaluate the efficiency of our method, we conducted an experiment comparing the total inference time between Mesh-Gait and DeepGaitV2~\cite{fan2023exploring}. For fair comparison, we employed 4DHuman~\cite{goel2023humans}, a robust algorithm for accurate 3D mesh inference, to generate 3D representations from video sequences. Specifically, we performed gait recognition on 6,000 image sequences with inferred 3D meshes. It is understood that different 3D reconstruction algorithms vary in inference speed, resulting in different efficiency comparisons with Mesh-Gait. This ablation study underscores the advantage of Mesh-Gait in real-world settings, as it eliminates the need for explicit 3D reconstruction from image sequences.

\section{Conclusion}

In this work, we introduce Mesh-Gait, a novel end-to-end multi-modal gait recognition framework that effectively integrates 2D silhouettes and 3D reconstructions to enhance gait recognition performance. Unlike traditional 2D-based methods, which often struggle with viewpoint variations, occlusion, and environmental noise, Mesh-Gait leverages 3D body shape information to generate a more robust and discriminative gait representation. Our approach efficiently reconstructs 3D representations directly from 2D silhouettes, eliminating the need for complex multi-view setups or expensive depth sensors, thereby significantly reducing system cost and complexity. Rather than directly learning point-based features from 3D joints or meshes which can be computationally expensive and difficult to integrate with silhouette-based features, Mesh-Gait learns an intermediate representation in the form of 3D heatmaps. These heatmaps capture rich spatial and structural information about the gait while providing a compact and efficient representation for fusion with silhouette features. Importantly, the reconstructed 3D heatmaps are progressively refined in a supervised manner using ground-truth 3D joints and meshes, ensuring precise spatial alignment and consistent 3D structure. By combining the complementary strengths of 2D silhouette features and intermediate 3D heatmap representations, Mesh-Gait significantly improves gait recognition accuracy while maintaining computational efficiency. This design makes real-time gait recognition feasible even in unconstrained environments. Through extensive evaluations on multiple benchmark datasets, Mesh-Gait consistently outperforms state-of-the-art methods in both accuracy and robustness, particularly under challenging conditions such as occlusion, large viewpoint changes, and noisy inputs, demonstrating its strong potential for real-world applications.

{\small
\bibliographystyle{ieee}
\bibliography{egbib}
}

\end{document}